\documentclass{article}

\usepackage{arxiv}

\usepackage[utf8]{inputenc} 
\usepackage[T1]{fontenc}    
\usepackage{hyperref}       
\usepackage{url}            
\usepackage{booktabs}       
\usepackage{amsfonts}       
\usepackage{nicefrac}       
\usepackage{microtype}      
\usepackage{lipsum}		
\usepackage{graphicx}
\usepackage{natbib}
\usepackage{doi}
\usepackage{subcaption}

\title{Performance in a dialectal profiling task of LLMs \\ for varieties of Brazilian Portuguese}

\author{ \href{https://orcid.org/0000-0002-4972-4320}{\includegraphics[scale=0.06]{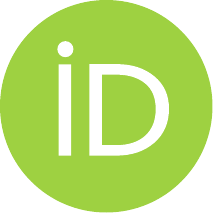}\hspace{1mm}Raquel Meister Ko Freitag}\thanks{We would like to express our sincere gratitude to the infrastructure and staff of the Multiuser Laboratory for Informatics and Linguistic Documentation (LAMID) for their invaluable support throughout this research.} \\
	Departamento de Letras Vernáculas \\ Universidade Federal de Sergipe -  UFS\\
	\texttt{rkofreitag@academico.ufs.br} \\
	\And
	\href{https://orcid.org/0009-0000-5270-8033}{\includegraphics[scale=0.06]{orcid.pdf}\hspace{1mm}Túlio Sousa de Gois} \\
	Departamento de Computação \\
    Universidade Federal de Sergipe - UFS\\
	\texttt{tuliosg@academico.ufs.br} \\
}

\renewcommand{\shorttitle}{Performance in a dialectal profiling task of LLMs for varieties of Brazilian Portuguese}

\begin{document}
\maketitle

\begin{abstract}
    Different of biases are reproduced in LLM-generated responses, including dialectal biases. A study based on prompt engineering was carried out to uncover how LLMs discriminate varieties of Brazilian Portuguese, specifically if sociolinguistic rules are taken into account in four LLMs: GPT 3.5, GPT-4o, Gemini, and Sabiá-2. The results offer sociolinguistic contributions for an equity fluent NLP technology.
\end{abstract}


\section{Introduction}
    Advances in generative AI have enabled near-human responses, crucial for overcoming the Turing test \cite{danziger2018intelligence}. However, achieving this requires algorithms to replicate ethically questionable human behaviors, including biases learned by large language models (LLMs) \cite{freitag2021preconceito}.
    
    Biases can be explicit, consciously manipulated, or implicit, operating unconsciously through automatic associations. These biases affect generative AI in two key areas: the rules and filters applied during LLM fine-tuning, and the linguistic datasets used for training. However, the specifics of these biases—whether in rules, filters, or dataset selection—remain unclear \cite{bender2021dangers}. To investigate these biases, reverse-engineering through prompt engineering is necessary, similar to how sociolinguistics studies human language attitudes.
    
    In Brazil, sociolinguistic studies over the past 50 years have highlighted significant asymmetries between prestigious and non-standard varieties (whether regional or socially stigmatized), often perpetuated by implicit biases in educational materials and media, such as the portrayal of regional accents in soap operas \cite{freitag2016sociolinguistica}. These societal biases likely extend to LLMs.

    For AI to be ethically and socially sensitive, the diversity of societal communities must be reflected in the language samples used to train LLMs. \cite{grieve2024sociolinguistic} define a language variety as ``a population of texts defined by external factors, such as the social background of the people who produce these texts, the social context in which these texts are produced, and the time period over which these texts are produced.''
    
    Currently, there is no transparency on how language samples are collected and balanced to reflect linguistic diversity. A study using prompt engineering could reveal how LLMs handle varieties of Brazilian Portuguese and whether they consider sociolinguistic rules.
    
\section{Dialectal biases in LLMs} \label{sec:firstpage}

Brazil's continental size contributes to its dialectal diversity, further enriched by social diversity in language use. Both geolinguistic and sociolinguistic approaches have systematically described these patterns in Brazilian Portuguese \cite{roncarati2003portugues, abraccado2015mapeamento}.

Though linguistic diversity doesn't align strictly with socio-political boundaries, it is socially perceived and manifests in stereotypes, such as the classic \textit{biscoito} vs. \textit{bolacha} 'cookie' debate,\footnote{https://rionoticias.com.br/afinal-bolacha-ou-biscoito/} regional jokes and memes,\footnote{https://www.conhecaminas.com/2016/10/10-coisas-que-so-mineiro-fala-entende-e.html} or even humorous maps that reflect aspects of perceptual dialectology \cite{preston2010language, freitag2015brasileiro,freitag2016brasileiros}.

\begin{figure}[h!]
     \centering
     \begin{subfigure}[b]{0.3\textwidth}
         \centering
         \includegraphics[width=\textwidth]{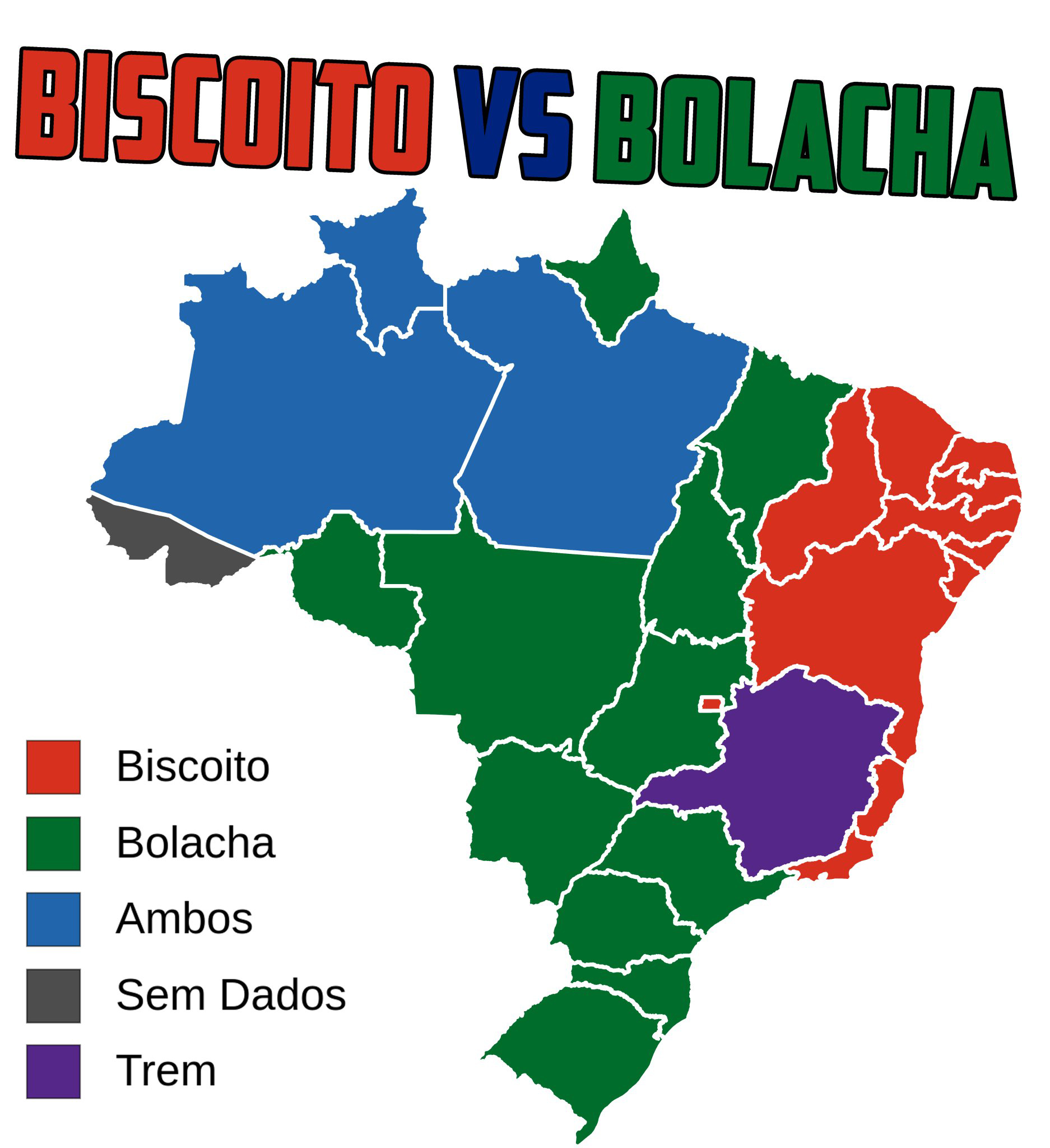}
         \caption{}
         \label{subfig:biscoito}
     \end{subfigure}
     \begin{subfigure}[b]{0.25\textwidth}
         \centering
         \includegraphics[width=\textwidth]{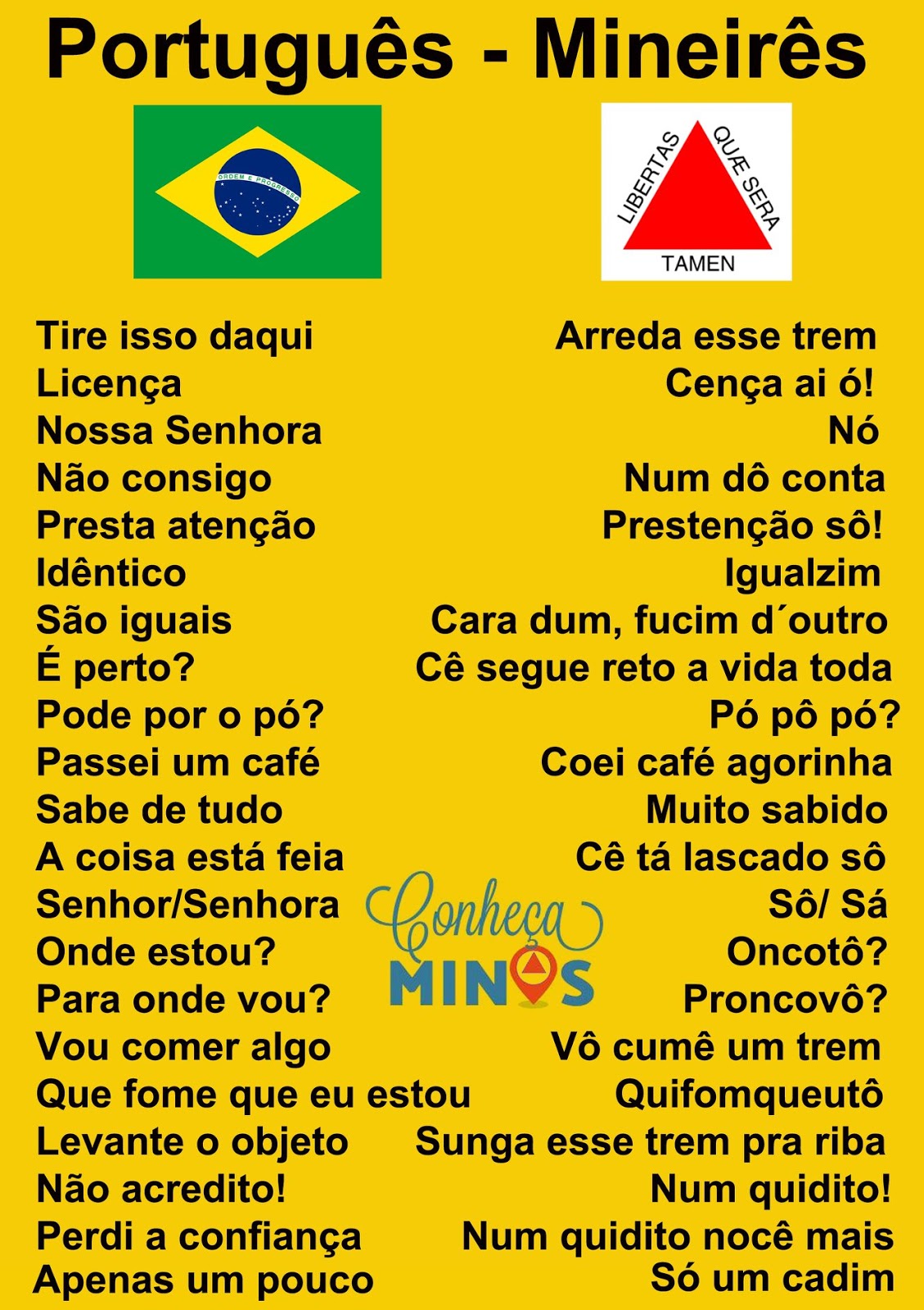}
         \caption{}
         \label{subfig:mineiros}
     \end{subfigure}
     \caption{\textit{Biscoito} or \textit{bolacha} (\subref{subfig:biscoito}) and mineiros's memes (\subref{subfig:mineiros})}
     \label{fig:jokes}
\end{figure}

\begin{figure}[h!]
     \centering
     \begin{subfigure}[b]{0.23\textwidth}
         \centering
         \includegraphics[width=\textwidth]{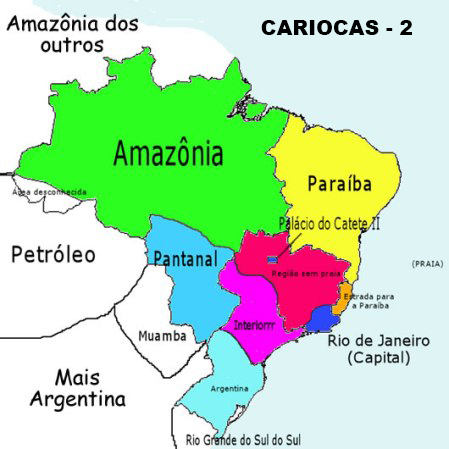}
         \caption{}
         \label{subfig:map1}
     \end{subfigure}
     \hfill
     \begin{subfigure}[b]{0.23\textwidth}
         \centering
         \includegraphics[width=\textwidth]{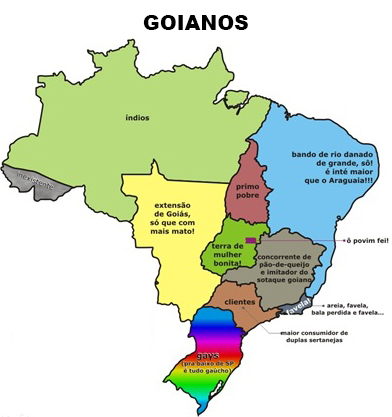}
         \caption{}
         \label{subfig:map2}
     \end{subfigure}
     \hfill
     \begin{subfigure}[b]{0.28\textwidth}
         \centering
         \includegraphics[width=\textwidth]{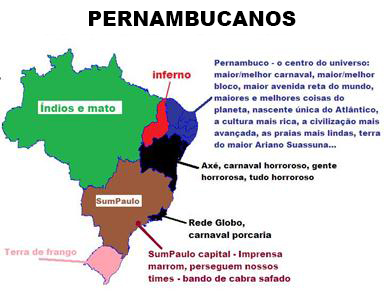}
         \caption{}
         \label{subfig:map3}
     \end{subfigure}
     \hfill
     \begin{subfigure}[b]{0.19\textwidth}
         \centering
         \includegraphics[width=\textwidth]{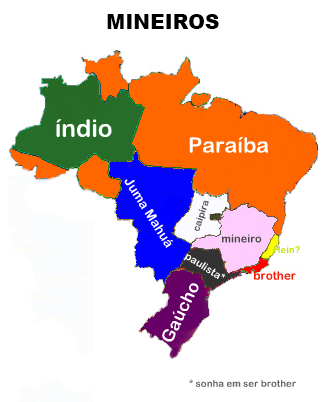}
         \caption{}
         \label{subfig:map4}
     \end{subfigure}
     \caption{Funny maps with Brazilian stereotypes}
     \label{fig:funny_maps}
\end{figure}

Sociolinguistic surveys support these representations, showing systematic patterns between dialectal groups and territories, such as variations in first-person plural (\textit{nós} ~ \textit{a gente}) and second-person pronouns (\textit{tu} ~ \textit{você}), their agreement (\textit{nós vamos} ~ \textit{nós vai} | \textit{tu vais} ~ \textit{tu vai}), and the shift between subjunctive and indicative moods for imperative forms (\textit{cante} ~ \textit{canta}) \cite{abraccado2015mapeamento}. Phonological variations, like diminutives and diphthongs, also appear in written representations, particularly in memes.

It is expected that LLMs, like humans, will learn and reproduce societal biases and stereotypes, including linguistic biases \cite{shrawgi2024uncovering, fleisig2024linguistic}, as observed in LLMs trained on African American English \cite{mengesha2021don, dacon2021truly, dacon2022evaluating, dacon2022towards}. This paper evaluates the accuracy of LLMs in identifying Brazilian Portuguese dialectal profiles, assessing both their agreement with human judgments and consistency across different models: GPT-4o and GPT 3.5 (OpenAI), Sabiá-2 (Maritaca AI), and Gemini (Google AI). None of these LLMs disclose the size of their training corpus or the sources of their texts.

\section{Method}
The procedures encompassed three stages: generation, classification, and data wrangling. The dataset and analysis scripts are available at \verb|https://osf.io/un8cw/.|

\subsection{Target-profiles generation}
The initial step involved instructing LLMs to generate text passages aimed at creating typical linguistic profiles for each of Brazil's 27 states. The prompt used was:

\textit{``Write a Facebook biography for my T-shirt store in XXX language, 
replacing XXX with the name of the state."}


Data was collected on 10 June 2024 and then analyzed for linguistic dialectal features to identify dialectal clues.

\subsection{Target-profiles classification}

LLMs were asked to identify the state of the text generated in the first step (\textit{task}) by two prompts: 
\begin{itemize}
\item \textit{task} + \textit{input}

\begin{quote}
\footnotesize
    \textit{``This is a target audience identification task. 
    Looking at the linguistic marks, identify for 
    target audience in which state of Brazil the 
    INPUT advertisement was constructed: \newline
    INPUT: XXXX"}
\end{quote}

\item \textit{task} + \textit{clue} based on  features identified by the human-judges + \textit{input}

\begin{quote}
\footnotesize
     \textit{``This is a target audience identification task. 
    Looking at the linguistic clues, identify for 
    the target audience in which state of Brazil the 
    has been constructed:\\
    INPUT: XXXX\\
    CLUES: interjections, second person pronoun, 
    first person plural pronoun, second person verb 
    agreement, first person plural verb agreement ''}
\end{quote}

\end{itemize}

\subsection{Data Wrangling}
 Data from the classification tasks was submitted to three stages of processing: 1) \textbf{Data cleaning}: city names were removed, retaining only state names; 2) \textbf{Standardization}: null answers were standardized as “não identificado,” while errors like “Sudoeste” and “Região Amazônica” were retained; and 3) \textbf{Data structuring}: all classification data was categorized and stored in \textit{.csv} files.

The models generated a total of 108 texts, 4 for each Brazilian state (1 per model) and 216 classifications were carried out, 108 for each type of prompt (\textit{task} and \textit{task + clue}). The generated texts were also classified by three human experts in syntactic dialectal variation, with the task of identifying the target state of the text. Measurement of the classifications (both of LLMs and humans) were carried out using Fleiss' Kappa \cite{fleiss1971measuring}.

\section{Results}

\subsection{Exploratory analysis of target-profiles}

The exploratory analysis revealed that Sabiá-2 \cite{sales2024sabia}, a Brazilian Portuguese-trained LLM from Maritalk, showed no dialectal variability in its responses. 

\begin{itemize}
    \item Sabiá-2 generation example:\vspace{1mm}
    \begin{quote}
    \footnotesize
    \textit{``Oi, [NOME DO LUGAR] Você já conhece a nossa loja de camisetas sustentáveis? Se ainda não, prepare-se para descobrir um novo jeito de expressar o seu amor pela nossa terra e pelo nosso planeta! Na nossa produção, usamos algodão orgânico, que respeita o ciclo da vida e a terra que pisa o nosso pé, e tintas naturais, que não sujam as águas que correm pelo nosso estado. As estampas são uma homenagem ao nosso estado: cada ponto turístico, cada paisagem natural, cada aspecto da nossa rica cultura é representado com orgulho e carinho. Você vai encontrar desde o estilo básico até o mais ousado, mas sempre com aquele toque eco-friendly que só nós sabemos dar. E aí, vai deixar de fazer parte dessa revolução verde? Na nossa loja, além de encontrar camisetas que fazem a diferença, você vai ter a certeza de que cada compra fortalece a nossa comunidade e a nossa luta por um mundo mais sustentável.Venha nos visitar, escolha a sua favorita e mostre que em [NOME DO LUGAR], além de um povo acolhedor e de uma natureza exuberante, temos a consciência ambiental como uma das nossas maiores riquezas! Com amor e orgulho [NOME DO LUGAR], [NOME DA LOJA]"}
    \end{quote}
    
\end{itemize}

In contrast, GPT 3.5, GPT-4o, and Gemini displayed sensitivity to dialectal differences, evident in variations of 2P pronouns (\textit{tu} ~ \textit{você}), 1PP agreement (\textit{nós vai} ~ \textit{nós vamos}), subjunctive/indicative morphology (\textit{compre} ~ \textit{compra}), interjections (\textit{uai, oxente, ôxe}), and discourse markers (\textit{visse? tá ligado? é nóix, bora lá}).

\begin{itemize}
    \item Gemini generation example: \vspace{1mm}
    \begin{quote}
    \footnotesize
    \textit{``E aí, pessoal! Bem-vindos à [NOME DA LOJA]! Aqui no coração da Amazônia, nóis tá fazendo moda de um jeito sustentável e cheio de orgulho da nossa terra. Nossas camisas são feitas com materiais reciclados e tinturas naturais, tudo bem cuidadoso pra respeitar o meio ambiente. O negócio é ficar estiloso e cuidar da nossa floresta ao mesmo tempo, visse? Aqui você encontra camisa que é a cara do [NOME DO LUGAR]: cheia de cor, vida e consciência. Vem com a gente nessa jornada de vestir a sustentabilidade com muito orgulho e amor pela natureza. Bora lá, junte-se a nós e venha conferir as novidades!"}
    \end{quote}
    
\end{itemize}

\begin{itemize}
    \item GPT 3.5 generation example: \vspace{1mm}
    \begin{quote}
    \footnotesize
    \textit{``E aí, meu chegado?! Seja bem-vindo à [NOME DA LOJA], a loja onde a moda sustentável é raiz! Aqui, o negócio é respeitar a floresta e o nosso mundão, com camisetas que são a cara do [NOME DO LUGAR]? Nossas camisas são tudo de bom, feitas com material reciclado e tingidas do jeito mais consciente, porque a gente se importa com a Amazônia e com a natureza que tá no nosso sangue, né não? Então, se tu quer se vestir com estilo e ainda ajudar a proteger o planeta, vem com a gente! Aqui tu vai encontrar o visual perfeito e ainda fazer bonito com o meio ambiente. Quando quiser, dá um pulo aqui na loja que a porta tá sempre aberta pra ti, meu amigo! Fica à vontade, viu?"}
    \end{quote}
\end{itemize}

\begin{itemize}
    \item GPT-4o generation example: \vspace{1mm}
    \begin{quote}
    \footnotesize
    \textit{``E aí, galera! Bem-vindos à [NOME DA LOJA]! Aqui é onde o estilo se encontra com o respeito pela natureza, do jeitinho que [NOME DO LUGAR] merece! Nós trabalhamos com material reciclado e tingimento responsável, tudo feito com carinho pras nossas terras [NOME DO LUGAR] e pro nosso planetão. É um compromisso sério, viu? Então, vem conferir nossas camisas que são um arraso, cheias de estilo e consciência ecológica. É moda que valoriza o que é nosso e cuida do futuro, firmeza? Bora junto nessa ideia de mostrar que em [NOME DO LUGAR], o verde tá na moda!"}
    \end{quote}
\end{itemize}

These features align with the descriptive patterns identified by previous sociolinguistic studies \cite{abraccado2015mapeamento}, reinforcing that some LLMs learn linguistic biases.

The generated responses were evaluated by three human judges specializing in syntactic dialectal variation. They identified the target state and provided reasoning for each input. Although their assessments highlighted similar dialectal clues as the exploratory analysis, the agreement was weak (inter-annotator: $\kappa = 0.31$; target: $\kappa = 0.13$). The next step was to have the LLMs perform the same task.

\subsection{Target-profiles evaluation}

\begin{figure}[h!]
\centering
\includegraphics[width=.8\textwidth]{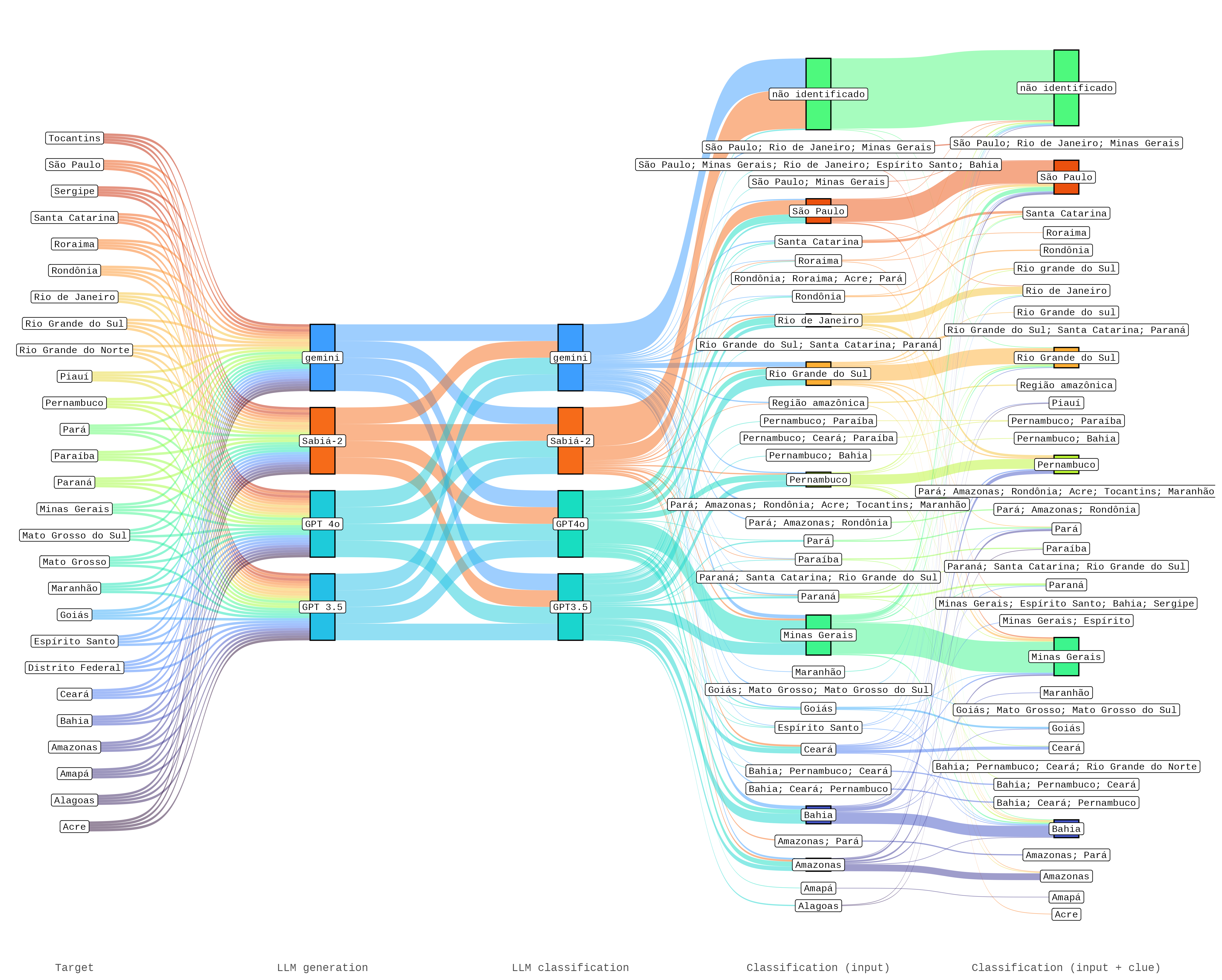}
\caption{Analysis flowchart}
\label{fig:flow2}
\end{figure}

LLMs were tasked with identifying the state of the text generated in the first step using two prompts: 1) \textit{task + input} and 2) \textit{task + clue + input}. The analysis flow is shown in Figure \ref{fig:flow2} and displays the Brazilian states, the LLMs used in both generation and classification, and the classifications performed with the two types of prompt. Although all prompts were in Brazilian Portuguese, Gemini responded in English.

In task 1, with weak agreement ($\kappa = 0.15$), differences emerged among LLMs: Sabiá-2 had the lowest agreement rate ($\kappa = 0.10$), while GPT-4o achieved the highest agreement, but still rather low ($\kappa = 0.21$). Identification rates varied as well: Sabiá-2 failed to identify a profile in 91\% of cases, followed by Gemini at 88\%, whereas GPT 3.5 and GPT-4o correctly identified the locations in over 20\% of cases. Unexpectedly, in task 2, the non-identification rate increased (not significant), suggesting that the linguistic features identified by humans did not significantly influence the LLMs' classification. The only exception was GPT 3.5, which correctly identified one more location (an increase of 0.92\%).

\begin{quote}
    
\begin{itemize}
    \item This is an example of Sabiá-2's answer for task 1: \newline
    \footnotesize
    \textit{``Observando as marcas linguísticas e o conteúdo da propaganda, podemos determinar que o público-alvo é de um estado brasileiro com características culturais e ambientais específicas, que valoriza a sustentabilidade e tem um modo de falar próprio. [...]
    Considerando todas essas informações, é possível inferir que o público-alvo da propaganda é de um estado do Brasil com características culturais e ambientais específicas, que valoriza a sustentabilidade e tem um modo de falar próprio. Sem uma especificação clara do \textit{[NOME DA LOJA]} ou do \textit{[NOME DO LUGAR]},  \textbf{não é possível determinar exatamente qual estado é o público-alvo}, mas as pistas sugerem um local com características semelhantes à  \textbf{região nordeste, dada a informalidade e o uso de gírias}, ou talvez  \textbf{um estado do sul ou sudeste que tenha uma comunidade engajada em questões ambientais e use redes sociais} como parte de sua estratégia de marketing."}
    
\end{itemize}
\end{quote}

The input lacks specific geographic identification; it could be attributed to the "Nordeste" region due to informal style and slang, or to the "Sul" or "Sudeste" regions, presumed to have higher engagement in social issues. GPT 3.5, while imprecise in geographic localization, provides a more detailed identification of grammatical features even without additional clues:

\begin{quote}
\footnotesize
\textit{``Com base nas marcas linguísticas presentes na propaganda do INPUT:
\begin{itemize}
    \item \textbf{Uso de "aí" e "tá ligado"}: Essas expressões são comuns em diversas regiões do Brasil, mas especialmente no contexto da propaganda, elas podem indicar uma influência mais forte de dialetos urbanos, como os encontrados em estados do Sudeste e Centro-Oeste.
    \item \textbf{Realização do pronome de primeira pessoa do plural}: "Nóis trabalha firme...", indicando um dialeto regional característico, comum em estados como São Paulo e Minas Gerais, por exemplo.
    \item \textbf{Expressões regionais como "mais da hora"}: Esta expressão é típica de regiões como o Sudeste e o Sul do Brasil.
    \item Referência ao compromisso com o planeta: A preocupação com sustentabilidade e meio ambiente é um tema geralmente valorizado em todo o Brasil, mas a forma como é apresentada na propaganda pode refletir uma sensibilidade urbana mais específica.
\end{itemize}
Com base nessas características linguísticas, a propaganda do INPUT parece ser direcionada principalmente para um público-alvo urbano, possivelmente das regiões \textbf{Sudeste ou Centro-Oeste do Brasil}, onde há uma mistura de expressões informais e um foco crescente em sustentabilidade."}
\end{quote}

GPT-4o provides the most accurate identification of the state intended by the input. The subsequent answer relates to the same input as those given by GPT 3.5 and Sabiá-2:

\begin{quote}
\footnotesize
\textit{``Observando as marcas linguísticas presentes na propaganda, podemos identificar o público-alvo provável com base nos seguintes elementos:
\begin{itemize}
    \item \textbf{Interjeições e expressões coloquiais}: "E aí", "tá ligado?", "nóis trabalha", "da hora", "bora", "sim senhor", "tá plantado".
    \item \textbf{Realização do pronome de segunda pessoa}: "você" não é explicitamente utilizado, mas a forma de tratamento é informal e coloquial.
    \item \textbf{Realização do pronome de primeira pessoa do plural}: "nóis" em vez de "nós".
    \item \textbf{Concordância verbal com segunda pessoa}: Não é explicitamente mostrada.
    \item \textbf{Concordância verbal com primeira pessoa do plural}: "nóis trabalha" em vez de "nós trabalhamos".
\end{itemize}
Essas marcas linguísticas são típicas do estado de São Paulo, especialmente da capital e regiões metropolitanas. O uso de "tá ligado?", "da hora" e \textit{"nóis"} são característicos do dialeto paulista, que muitas vezes é associado à \textit{cultura urbana paulistana}.
Portanto, a propaganda do INPUT parece ser construída para o público-alvo do \textbf{estado de São Paulo}."}
\end{quote}

GPT-4o, considering all the grammatical features, determines that the input is intended for São Paulo, providing a more precise answer than GPT 3.5. Like humans, GPT-4o integrates linguistic clues and covariation, as illustrated in the following example:

\begin{quote}
\footnotesize
\textit{``A análise das marcas linguísticas presentes no texto da propaganda do INPUT sugere que o público-alvo é o estado de Pernambuco. Aqui estão as razões:
\begin{enumerate}
    \item \textbf{Interjeições e expressões regionais}: A expressão "visse?" é tipicamente pernambucana e nordestina.
    \item \textbf{Realização do pronome de segunda pessoa}: A forma "você" é usada, que é comum em várias regiões do Brasil.
    \item \textbf{Realização do pronome de primeira pessoa do plural}: "Nóis" em vez de "nós" indica uma variação regional.
    \item \textbf{Concordância verbal com segunda pessoa}: "deixar você", "ajudar nosso planeta".
    \item \textbf{Concordância verbal com primeira pessoa do plural}: "Nóis trabalha", uma forma de falar mais informal e regional.
\end{enumerate}
Outras expressões como "galera", "arretadas de boas" e "bora" reforçam ainda mais o tom informal e típico da linguagem usada no estado de Pernambuco.
Portanto, a propaganda do INPUT foi construída para o público-alvo de Pernambuco, usando elementos linguísticos característicos da região para criar uma conexão mais forte com os consumidores locais."}
\end{quote}

The feature associated with urbanity in the previous answer is seen as a regional trait in this case, leading to the attribution of the input to Pernambuco. This variation can be explained by indexical fields \cite{eckert2008variation}. For instance, \textit{nós vai} is considered an informal feature for Pernambuco but an indicator of urban culture for São Paulo. This pattern aligns with societal studies of Brazilian Portuguese \cite{freitag2015brasileiro}.

\begin{figure}[h!]
\centering
\includegraphics[width=.8\textwidth]{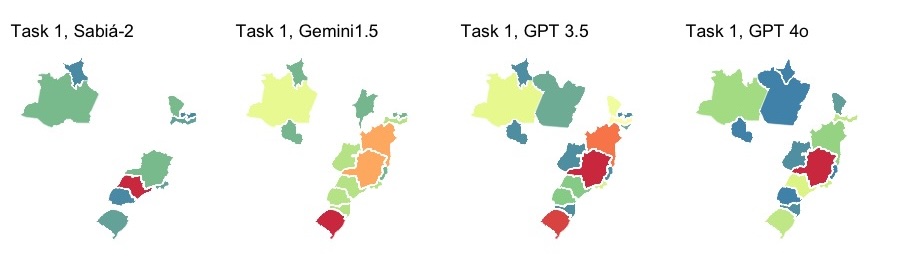}
\caption{Geographical distribution of profile identification in task 1 \textit{prompt: task + input}}
\label{fig:task1}
\end{figure}

\begin{figure}[h!]
\centering
\includegraphics[width=.8\textwidth]{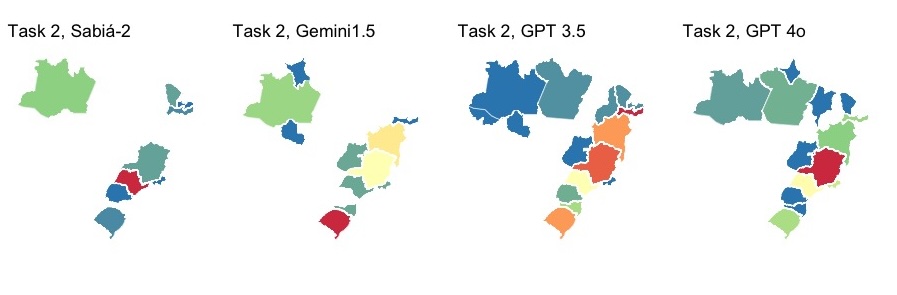}
\caption{Geographical distribution of profile identification in task 2 \textit{prompt: task + clue + input}}
\label{fig:task2}
\end{figure}

In task 1), the most common states identified were Minas Gerais (9\% for GPT-4o and 4\% for GPT 3.5), São Paulo (5\% for Sabiá-2), and Rio Grande do Sul (2\% for Gemini) (Figure \ref{fig:task1}). In task 2), results were similar except GPT 3.5, which identified Pernambuco (Figure \ref{fig:task2}). Only GPT 3.5 and GPT-4o provided explicit analyses of clues, such as \textit{tu} being common in Southern Brazil or \textit{nóis vai} indicating informality or urban features. These findings highlight the sociolinguistic fine-tuning of LLMs or their \textit{language regard} \cite{preston2010language}.

\section{Discussion}

This study investigated dialectal sensitivity in LLMs by assessing their responses to tasks aimed at identifying dialectal features in Brazilian Portuguese.

The findings revealed that Sabiá-2, a chatbot from Maritalk trained in Brazilian Portuguese, lacked dialectal sensitivity, showing no variability in responses. In contrast, GPT 3.5, GPT-4o, and Gemini demonstrated sensitivity to dialectal differences, evident in variations in pronoun usage, verb agreement, and other linguistic markers, aligning with sociolinguistic patterns. The agreement among human evaluators on dialectal features was weak ($\kappa = 0.31 $), suggesting either inconsistent identification or a low number of judges. Among LLMs, Sabiá-2 had the lowest agreement rate ($\kappa = 0.08$), while GPT-4o showed the best agreement among the classifications ($\kappa = 0.21$). Notably, Gemini responded in English despite prompts being in Portuguese.

Incorporating specific linguistic features into prompts did not notably improve the LLMs' ability to identify the state, indicating that these features may not significantly affect classification. In task 1), GPT-4o and GPT 3.5 often identified Minas Gerais, while São Paulo and Rio Grande do Sul were identified by Sabiá-2 and Gemini1.5, respectively. In task 2), GPT 3.5 shifted its identification to Pernambuco. Only GPT 3.5 and GPT-4o provided explicit justifications based on dialectal clues, indicating some understanding of regional features.

These results show that while LLMs can detect dialectal variation, their ability to pinpoint specific regional profiles is inconsistent. The use of human-identified linguistic clues does not significantly enhance classification accuracy. Understanding how LLMs handle language varieties can help sociolinguistics explore human processing of linguistic variation and contribute to advancing linguistic justice and equitable NLP technologies \cite{baugh2018linguistics, wolfram2019implicit, nee2021advancing, nee2022linguistic, liu2023dada}.

\bibliographystyle{unsrtnat}
\bibliography{references}

\end{document}